# BF1: A Causal Dyadic Sparse-Attention Retrofit for Efficient Long-Context Transformers

*Correctness, Systems Scaling, and Replicated Matched Adaptation*

**Hina Dixit**

Decompute Inc., Cupertino, CA, USA

August 2026

## ABSTRACT

Dense causal attention remains expensive at long context even when implemented with highly optimized exact kernels. We study BF1, a deterministic block-aligned dyadic sparse-attention route that combines a small exact local neighborhood, a global first block, and logarithmically spaced historical blocks. The route is related to prior log-sparse and dilated attention patterns; our contribution is a correctness-gated pretrained-model retrofit, a matched topology-control study, and a systems characterization that connects per-layer sparsity to whole-model latency. For fixed block width, every converted layer uses O(n log n) selected token interactions and has O(log n) graph communication depth. On an NVIDIA RTX PRO 6000 Blackwell GPU, an optimized BF16 implementation crosses dense attention between 2K and 4K tokens and reaches a 10.91× per-layer prefill speedup at 32K. Retrofitting eight of 28 Qwen3-0.6B attention layers lowers warm whole-model time to first token by 7.7%, 11.3%, and 15.3% at 8K, 16K, and 32K, respectively, while the remaining dense layers keep the complete model asymptotically quadratic. Under a matched 1,000-step, 16.384M-token adaptation protocol, BF1 ranks first across three training seeds: mean report perplexity is 1.68639 versus 1.69154 for a matched static-random nonlocal graph, 1.69258 for dense continued training, and 1.81505 for equal-budget local sliding. At seed 1234, the packed-report paired interval places Dense-CT 0.3169–0.4055% above BF1 and static-random graph 17 0.2441–0.3642% above BF1. These results establish BF1 as a sparse operator and selective retrofit primitive, replicated across three tested training seeds, with real long-context systems value. This paper evaluates numerical correctness, selected-interaction scaling, kernel performance, partial-model inference, and matched next-token language modeling; it does not claim a fully subquadratic model, compressed total KV storage, or general preservation of retrieval, aggregation, and state-tracking capability.

## 1 Introduction

Transformer self-attention gives each query direct access to all earlier key–value pairs, but its causal interaction count grows quadratically with sequence length (Vaswani et al. 2017). IO-aware exact kernels such as FlashAttention greatly reduce memory traffic without changing the all-pairs interaction law (Dao, Fu, et al. 2022). A long line of work therefore replaces dense attention with fixed sparse patterns, dilated routes, hierarchical approximations, or query-adaptive selection (Child et al. 2019; Li et al. 2019; Beltagy et al. 2020; Zaheer et al. 2020; Ding et al. 2023; Tang et al. 2024; Yuan et al. 2025).

This paper studies a narrower engineering question: can a deterministic causal multiscale route be retrofitted into a pretrained language model, trained under matched controls, and shown to cross optimized dense attention in wall-clock latency? We call the resulting operator BF1. The public route is block aligned and data independent: each query block reads itself causally, a small local neighborhood, the first block, and predecessor blocks at dyadic distances. BF1 is an internal name retained for continuity; mathematically, the operator is a causal dyadic/log-sparse block graph rather than a claim that exponentially spaced predecessors are new.

The systems result is positive. BF1 crosses dense attention at 4K in the tested Qwen3-0.6B geometry and reaches 10.91× per-layer speedup at 32K. The selected eight-layer retrofit lowers warm 32K whole-model TTFT by 15.3%. The language-modeling result also replicates: BF1 ranks first in all three matched Stage A training runs against dense continued training, a fixed random nonlocal graph, and equal-budget local sliding.

The paper makes five contributions:

- We specify a reproducible causal dyadic block route and prove its O(n log n) selected-interaction law and logarithmic communication depth per converted layer.
- We provide correctness-gated BF16 Blackwell measurements, including partial blocks, backward gradients, dense-path equivalence, future-token isolation, and opposite-order benchmarks through 32K.
- We report whole-model TTFT for the actual eight-layer retrofit and separate it from an all-layer timing-only ceiling.
- We conduct a three-training-seed matched adaptation study with Dense-CT, equal-pair-count local sliding, and an equal-pair-count static-random nonlocal graph.
- We characterize selected-page decode semantics and show that planning lifecycle, rather than selected-page attention arithmetic, is the dominant unresolved low-batch cost.

Our scope is intentionally bounded. The present result establishes an operator, a systems regime, and matched next-token language-modeling behavior. Capability-level retrieval, aggregation, and mutable-state analyses are outside the reporting scope of this paper and are addressed separately.

## 2 Related Work

**Fixed and dilated sparse routes.** Sparse Transformer, Longformer, BigBird, and Reformer reduce attention cost through factorized, local/global, random, or hashing-based patterns (Child et al. 2019; Beltagy et al. 2020; Zaheer et al. 2020; Kitaev et al. 2020). LogSparse Transformer uses logarithmically spaced predecessors and local enhancements, making it the closest early route-level predecessor to BF1 (Li et al. 2019). LongNet likewise uses exponentially dilated attention and emphasizes logarithmic dependency between tokens (Ding et al. 2023). H-Transformer-1D applies hierarchical matrix structure to long-range attention (Zhu and Soricut 2021), while Pixelated Butterfly applies butterfly-structured sparsity more broadly to neural network matrices (Dao,

Chen, et al. 2021). BF1 does not claim novelty for dyadic predecessors alone. Its contribution is the block-aligned pretrained-model retrofit, matched topology controls, Blackwell kernel study, and whole-model attribution.

**Adaptive long-context inference.** Quest selects KV-cache pages using the current query (Tang et al. 2024); DuoAttention allocates full context to retrieval heads and bounded caches to streaming heads (Xiao et al. 2024); HiP uses hierarchical pruning without retraining (Lee et al. 2025); Native Sparse Attention combines compression, selection, and local context in a natively trained architecture (Yuan et al. 2025); and XAttention scores sparse blocks through antidiagonal structure (Xu et al. 2025). These approaches address query specificity or head heterogeneity that a deterministic route intentionally does not model.

**Sparse-attention systems.** FlashAttention establishes the importance of IO-aware exact kernels (Dao, Fu, et al. 2022). Sparse Flash Attention extends flash-style execution to broad sparsity patterns (Pagliardini et al. 2023). BF1 uses FlexAttention as a compiler interface for custom block masks (Dong et al. 2024) and evaluates exact selected physical pages using FlashInfer-style paged attention semantics (Ye et al. 2025). We report pair reduction, sparse-kernel latency, planning cost, and whole-model TTFT separately.

## 3 BF1 Operator

### 3.1 Causal dyadic block graph

Let a sequence of n tokens be partitioned into $B = \lceil n/b \rceil$ blocks of width b, indexed from 0 to $B - 1$. For query block j, define

$$P(j) = \{j, j - 1, j - 2, 0\} \cup \{j - 2^r : r \geq 2\}, \tag{1}$$

with invalid and duplicate indices removed. The self block uses a token-level causal triangle; all other selected blocks are strictly earlier. Figure 1 shows a representative route. The reported configuration uses $b = 64$, two local predecessor blocks, the global first block, and dyadic offsets beginning at four blocks. No scale-dependent score bias is used.

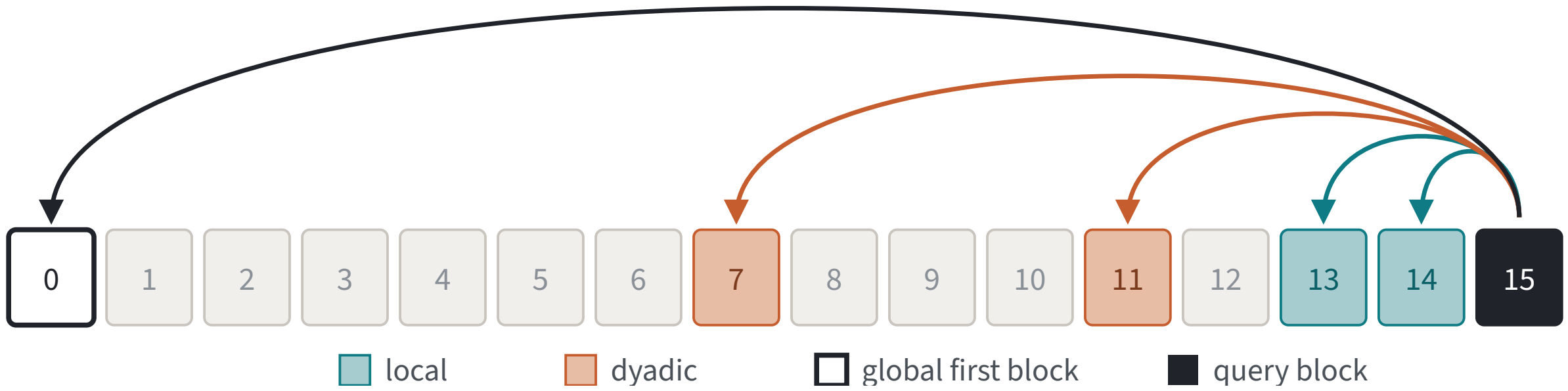

**Figure 1: BF1 block route.** A query block receives exact local context, the first global block, and dyadic historical predecessors. The public route is fixed, causal, and block aligned.

### 3.2 Selected-interaction complexity

**Proposition 1** *(Per-layer selected interactions). For fixed block width b, BF1 uses $O(n \log n)$ selected token interactions in a causal attention layer.*

*Proof.* Each block j selects at most $4 + \lfloor \log_2 j \rfloor$ predecessor blocks. Thus the number of selected block pairs is

$$\sum_{j=0}^{B-1} O(1 + \log(j + 1)) = O(B \log B).$$

Each selected block pair contributes at most $b^2$ token pairs, giving $O(b^2 B \log B) = O(nb \log(n/b))$. For fixed b, this is $O(n \log n)$. The causal triangle inside the self block only lowers the count. □

For one-token cached decode, a query reads $O(\log B)$ selected blocks and therefore $O(bd_h \log B)$ key/value elements per head. The underlying exact historical KV store remains $O(n)$; BF1 is not a logarithmic-memory method.

### 3.3 Communication depth

For non-global blocks $i < j$, write $\delta = j - i$ in binary. Repeatedly subtracting the largest present power of two gives a path whose length is at most the population count of $\delta$; local offsets handle the two least significant distances. The global first block is directly visible.

**Proposition 2** *(Dyadic reachability). For B blocks, every earlier block can influence a later block in O(log B) BF1 layers. In the evaluated B = 512 graph, the maximum shortest path is eight layers.*

*Proof.* If $i = 0$, the global edge makes block 0 directly visible to every later block, giving a one-layer path. If $\delta = j - i \in \{1, 2\}$, the local edges to $j - 1$ and $j - 2$ give a one-layer path. Otherwise, decompose $\delta$ into distinct powers of two, $\delta = 2^{r_1} + \cdots + 2^{r_m}$ with $r_1 > \cdots > r_m$. Starting from i and adding the powers in decreasing order produces monotonically increasing intermediate positions $i < i + 2^{r_1} < \cdots < j$, each at most j and therefore a valid block index. Every step of size $2^r$ with $r \geq 2$ is a dyadic predecessor edge under Equation 1, and steps of size one or two are local edges, so each consecutive pair on the path is connected in one BF1 layer. The path length is therefore at most $\mathrm{popcount}(\delta) \leq \lfloor \log_2 B \rfloor$. For $B = 512$ this bound gives nine layers; the only distance with nine one-bits is $\delta = 511$, which forces $i = 0$ and is served by the direct global edge, so the finite-graph maximum shortest path is eight layers. □

This is a communication-graph statement. It characterizes structural reach under the declared route; it does not by itself establish any downstream task capability.

### 3.4 Whole-model asymptotic scope

The evaluated Qwen retrofit converts eight of 28 attention layers and leaves 20 layers globally dense.

**Proposition 3** *(Mixed-stack complexity). If a fixed-depth model contains at least one unrestricted dense global-attention layer, its sequence-length attention complexity remains $\Omega(n^2)$.*

*Proof.* With $L_d$ dense layers and $L_s$ BF1 layers,

$$T(n) = \Theta(L_d n^2 d_h) + O(L_s n \log n \, d_h) + T_{other}(n).$$

For any constant $L_d \geq 1$, the dense term is $\Omega(n^2)$. □

The current model therefore provides a measured constant-factor acceleration rather than a whole-model O(n log n) claim.

# 4 Experimental Method

## 4.1 Hardware and software

Experiments use an NVIDIA RTX PRO 6000 Blackwell Workstation Edition GPU (compute capability 12.0, approximately 95 GiB usable memory), PyTorch 2.13.0 with CUDA 13.0, BF16 attention, and Qwen3-0.6B-Base (Yang et al. 2025). The target geometry has 28 decoder layers, 16 query heads, eight KV heads, and head dimension 128.

## 4.2 Correctness gates

Performance and model results are interpreted only after the following checks pass:

- exact route and selected-pair accounting;
- zero output change under mutation of future tokens;
- BF16 agreement with a materialized routed-mask reference;
- compiled forward and backward with finite Q/K/V gradients;
- partial final-block coverage at length 257;
- dense-path equivalence when the sparse blend is closed;
- cached one-token decode and independent blockwise reference agreement.

The target FlexAttention cases at lengths 128, 257, and 512 have maximum absolute error 0.00390625 and mean error below $7.1 \times 10^{-5}$ in BF16, with zero measured future leakage.

## 4.3 Performance protocol

Prefill measurements use batch size one, five warmups, 30 measured repetitions, and two op posite pattern orders. Reported values are the median of the two order medians. Dense native SDPA and compiled BF1 use the same attention geometry. Equal-pair-count sliding and static-random routes are timed separately to test whether topology changes physical sparse-kernel cost.

Whole-model TTFT is measured in a common eager Hugging Face harness at 8K, 16K, and 32K for: (i) the selected checkpoint on the dense path; (ii) the same checkpoint with eight Stage A BF1 layers active; and (iii) an all-layer BF1 timing-only configuration. The third condition is a systems ceiling, not a trained model result.

### 4.4 Matched Stage A adaptation

Students and a frozen dense teacher are initialized from Qwen3-0.6B-Base. The same eight middle attention layers are adapted in every arm. BF1, matched sliding, and static random progressively activate sparse routes; Dense-CT retains the same retrofit and trainable attention parameters but keeps the sparse blend at zero. The objective combines next-token loss, teacher-logit distillation, and hidden-state matching.

A 49,000-row ORCA-style question–answer corpus (Mukherjee et al. 2023) is normalized and split by content hash into train, select, and report sets. Each run consumes 16.384M training tokens over 1,000 optimizer steps. Checkpoints are selected by minimum held-out select NLL before one-time report evaluation. Every report uses the same ordered 174 packed sequences and 356,178 predicted tokens. The four-arm campaign is repeated at training seeds 1234, 2026, and 3407. Static-random graph seed 17 is fixed to isolate optimizer/training-seed variation.

**Table 1: Matched Stage A arms.** All rows use 50,333,696 trainable parameters, the same token order, objective, optimizer schedule, checkpoint cadence, held-out selection rule, and report packs. Sparse arms also use the same selected-pair budget.

| Arm | Active path | Route | Blend |
|---|---|---|---:|
| BF1 | sparse | causal dyadic | 1 |
| Random-17 | sparse | static nonlocal | 1 |
| Sliding | sparse | contiguous local | 1 |
| Dense-CT | dense | sparse inactive | 0 |

### 4.5 Evaluation scope

This paper evaluates numerical correctness, route complexity, kernel scaling, partial-model latency, and matched next-token perplexity. It does not report a general capability evaluation and makes no claim about preservation of retrieval, multi-source aggregation, mutable-state tracking, or arbitrary long-context reasoning.

# 5 Systems Results

## 5.1 Per-layer prefill scaling

Figure 2 reports the optimized BF1 operator. Dense attention is faster at short context, where sparse metadata and launch overhead dominate. BF1 crosses dense between 2K and 4K, reaches 2.84× at 8K, 5.50× at 16K, and 10.91× at 32K. The 32K selected-interaction count falls by 26.98×, so the implementation realizes a substantial but incomplete fraction of the pair-count reduction.

Equal-pair-count BF1, sliding, and static-random routes have nearly identical sparse kernel time. BF1's distinction at matched physical cost is therefore its communication-depth topology, not cheaper per-layer arithmetic.

**Table 2: Per-layer BF16 prefill speedup versus dense SDPA.** Values as reported in the text; BF1 crosses dense between 2K and 4K.

| Context | Measured speedup | Interaction reduction |
|---|---:|---:|
| 8K | 2.84× | — |
| 16K | 5.50× | — |
| 32K | **10.91×** | 26.98× |

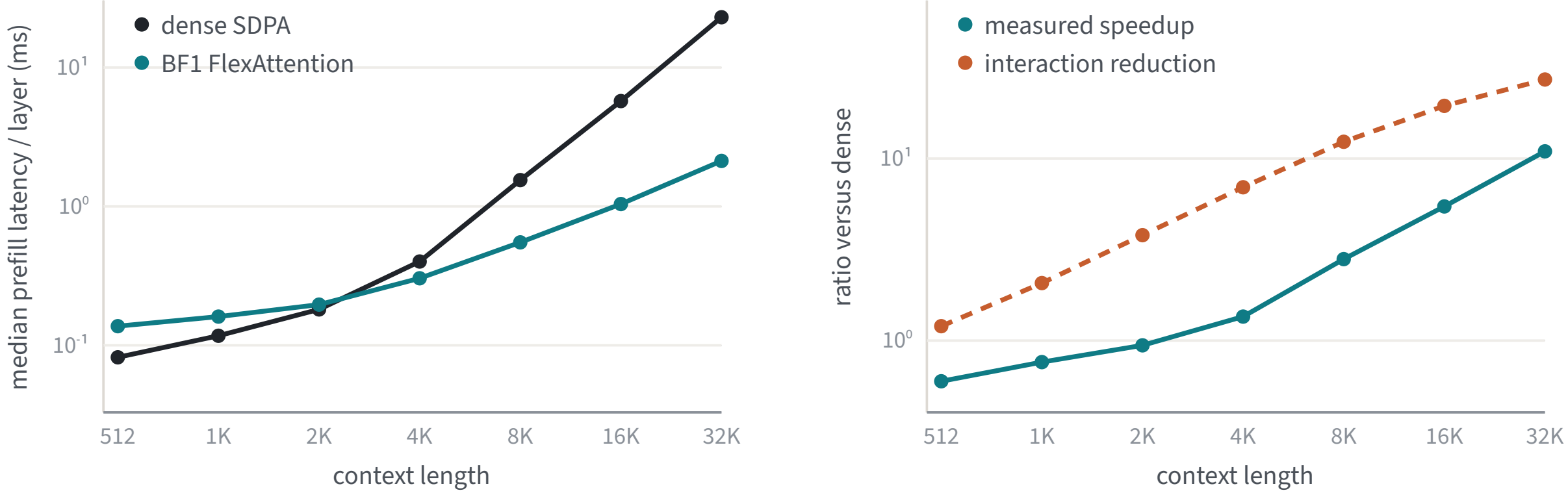

**Figure 2: Order-controlled per-layer BF16 prefill.** Left: dense and BF1 latency. Right: measured speedup and selected-interaction reduction.

### 5.2 Communication depth

Figure 3 compares maximum shortest paths. At 32K, BF1 requires eight hops, static-random graph 17 requires 14, and matched sliding requires 59 under the same selected-pair budget.

The graph metric explains why local sliding is structurally different from the two nonlocal sparse controls. It does not imply that shortest path alone determines language-modeling quality.

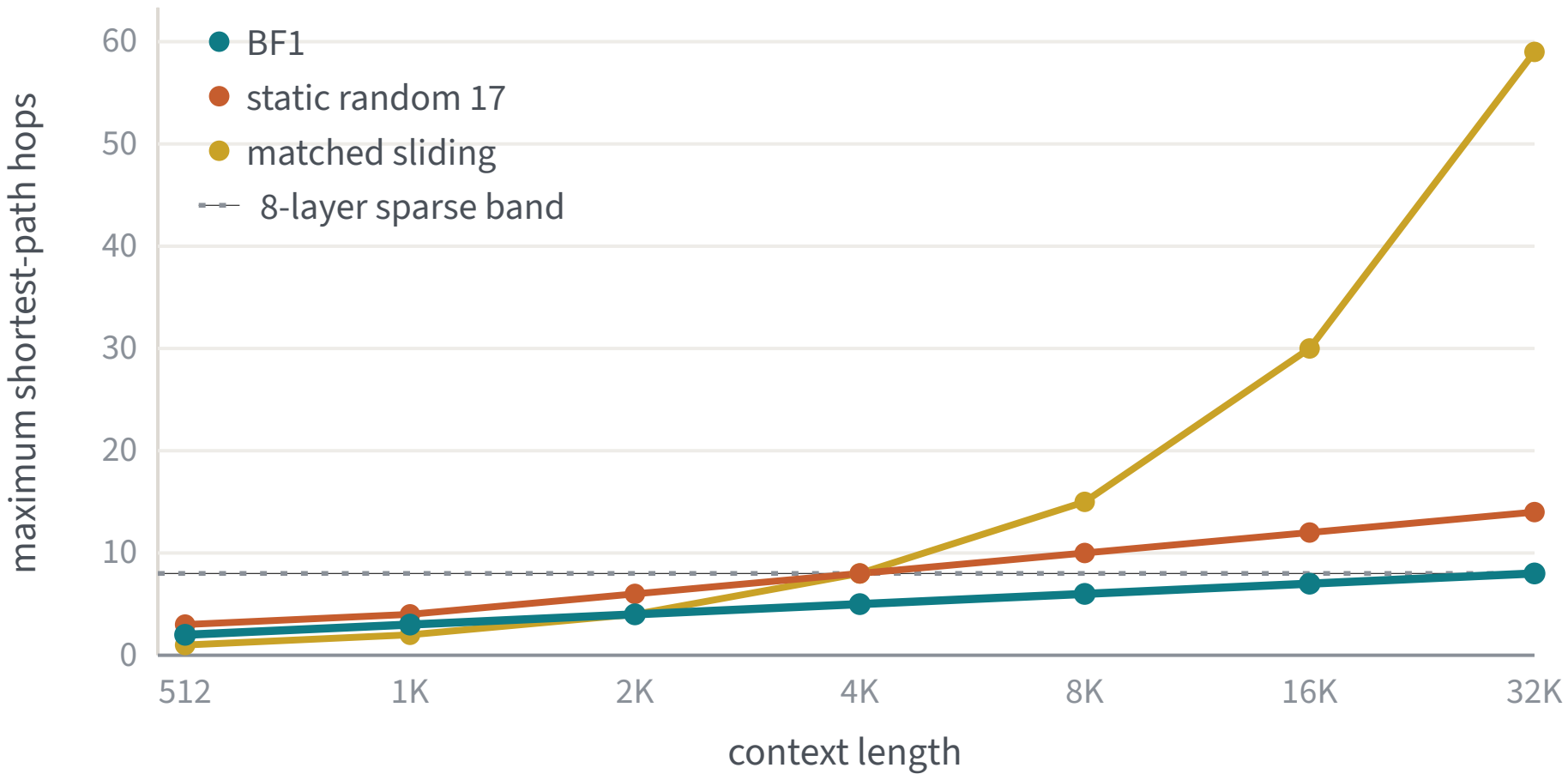

**Figure 3: Maximum shortest-path communication depth at matched sparse interaction budget.** The dotted line marks the eight-layer sparse band.

### 5.3 Whole-model TTFT

The actual eight-layer retrofit lowers warm TTFT from 69.0 to 63.7 ms at 8K, 190.1 to 168.6 ms at 16K, and 589.1 to 499.1 ms at 32K, corresponding to 1.08×, 1.13×, and 1.18× speedups. A second session gives a 15.7% 32K reduction, consistent with the primary 15.3% result. Peak VRAM is unchanged in this harness.

The all-layer timing-only condition reaches 2.14× at 32K. Its near-additive scaling with the converted-layer count is a useful systems ceiling, but it is not a trained all-layer result. An additional kernel sweep improved 32K per-layer BF1 time by about 14% (from 9.56× to 10.91× versus dense) while changing the eight-layer whole-model TTFT by less than 0.5%. This is an Amdahl-style result: after only eight layers are converted, safe coverage of more layers can matter more than further tuning of an already-fast operator.

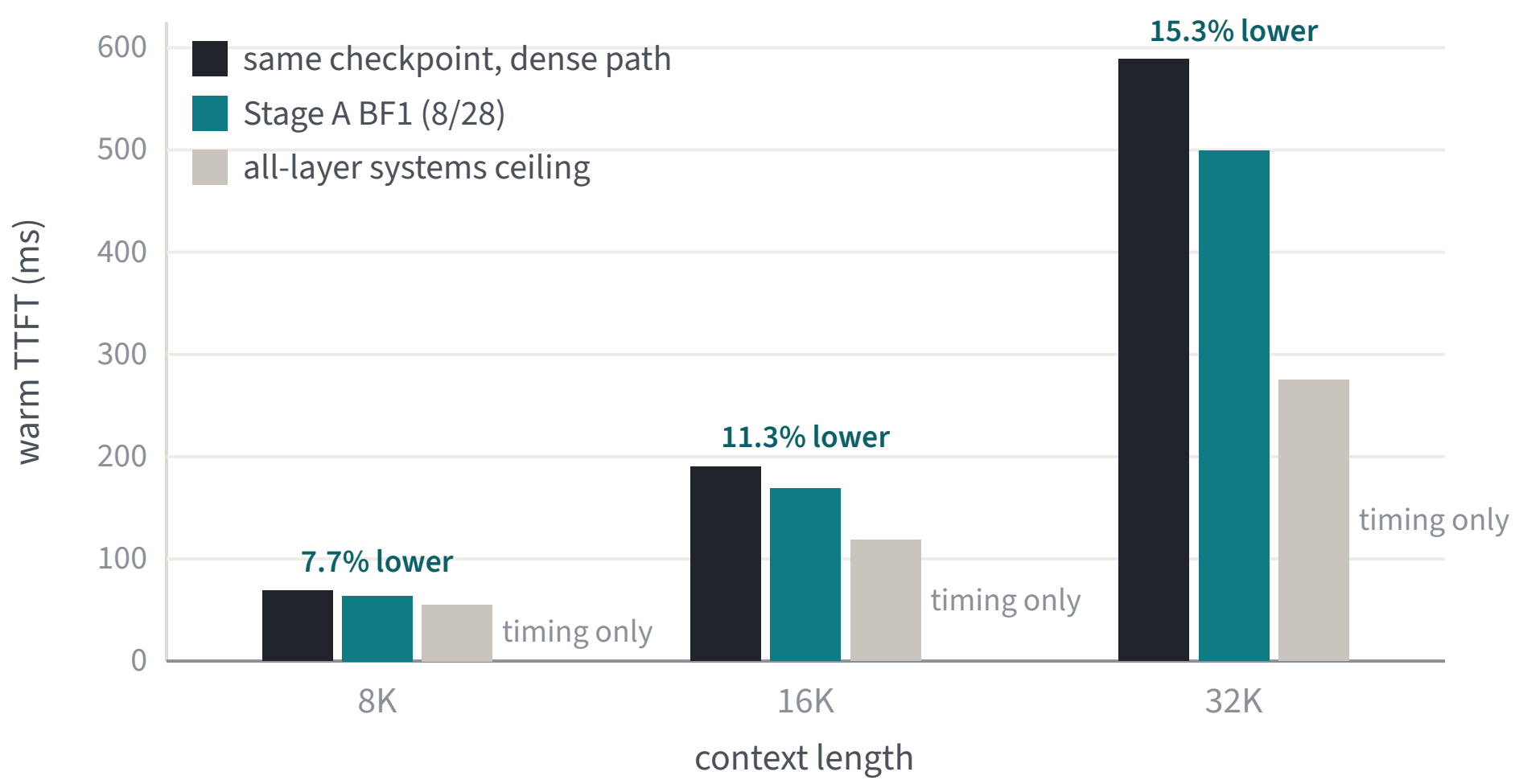

**Figure 4: Warm whole-model TTFT in the common eager harness.** The gray all-layer condition is a timing ceiling only.

### 5.4 Selected-page decode characterization

The correctness-oriented gathered decode path selects BF1 tokens, materializes compact K/V tensors, and calls native SDPA. It reaches approximately 1.21× versus dense at 32K, batch one, while retaining the full O(n) KV cache. A paged-GQA backend executes the exact selected-page attention with BF16 numerical agreement, tail handling, page-permutation invariance, and zero decoy-page leakage.

Subtracting a separately measured plan charge from the effective call yields a residual execution estimate of 12–23 microseconds, approximately 6–18× below gather across the tested batch range. Charging a 100–121 microsecond plan on every token, however, makes the effective low-batch path slower; batch 32 reaches 2.67× at 32K and 3.05× at 64K. The remaining decode systems problem is persistent plan reuse across a 64-token route epoch. We do not claim a low-batch end-to-end decode win.

**Table 3: Selected-page paged-GQA attention-call decomposition.** Times are microseconds. "Residual execution" is effective latency minus the separately measured plan charge; it is an estimate rather than a direct persistent-plan measurement.

| Context | Batch | Gather | Effective | Plan | Residual execution | Effective speedup |
|---|---|---|---|---|---|---|
| 32K | 1 | 76.112 | 114.831 | 102.855 | 11.976 | 0.66× |
| 32K | 8 | 100.584 | 116.471 | 102.855 | 13.616 | 0.86× |
| 32K | 32 | 382.280 | 143.064 | 121.352 | 21.712 | **2.67×** |
| 64K | 1 | 76.568 | 112.781 | 100.381 | 12.400 | 0.68× |
| 64K | 8 | 105.560 | 115.922 | 102.274 | 13.648 | 0.91× |
| 64K | 32 | 419.728 | 137.606 | 114.622 | 22.984 | **3.05×** |

## 6 Matched Language-Modeling Results

Table 4 and Figure 5 show the frozen report results. BF1 ranks first in all three training runs. Dense-CT is 0.358–0.384% above BF1; static-random graph 17 is 0.300–0.313% above BF1; matched sliding is 7.623–7.637% above BF1. The magnitudes are stable across training seeds.

The seed-1234 packed-report intervals provide a within-report uncertainty audit. Relative to BF1, static random is +0.3030% with a 95% paired interval of [+0.2441, +0.3642]%; Dense-CT is +0.3599% with [+0.3169, +0.4055]%; and matched sliding is +7.6367% with [+7.3325, +7.9554]%. These are report-pack intervals for one frozen model pair, not intervals over independent training runs.

The unadapted dense base has report perplexity 1.77161. Dense-CT recovers approximately 92.7% of the absolute perplexity reduction from the unadapted base to BF1 (approximately 92.6% when measured in NLL, the additive quantity), so continued in-domain adaptation is the dominant source of the gain. The residual BF1 increment nevertheless replicates in 3/3 training runs.

The topology controls support two bounded conclusions:

1. nonlocal sparse connectivity is load-bearing under this protocol, because equal-budget local sliding is much worse;
2. BF1 consistently beats static-random graph 17, but graph-draw replication is required before claiming a distribution-level advantage over random sparse graphs.

**Table 4: Report perplexity across training seeds.** Lower is better. The same report corpus is reused across seeds, so the 174 report packs are not pooled as 522 independent observations.

| Arm | Seed 1234 | Seed 2026 | Seed 3407 | Mean | Sample SD |
|---|---:|---:|---:|---:|---:|
| BF1 dyadic | **1.68634** | **1.68622** | **1.68660** | **1.68639** | 0.00019 |
| Static random, graph 17 | 1.69145 | 1.69128 | 1.69188 | 1.69154 | 0.00031 |
| Dense-CT, blend 0 | 1.69241 | 1.69226 | 1.69308 | 1.69258 | 0.00044 |
| Matched sliding | 1.81512 | 1.81486 | 1.81517 | 1.81505 | 0.00017 |

**Table 5: Seed-1234 paired bootstrap audit over the fixed ordered report packs.** Relative change is the other arm's perplexity versus the reference.

| Reference | Other | Relative PPL change | 95% paired interval |
|---|---|---:|---:|
| Unadapted dense | BF1 | −4.8129% | [−4.9669, −4.6561]% |
| Unadapted dense | Dense-CT | −4.4703% | [−4.6485, −4.2917]% |
| BF1 | Static random 17 | +0.3030% | [+0.2441, +0.3642]% |
| BF1 | Dense-CT | +0.3599% | [+0.3169, +0.4055]% |
| BF1 | Matched sliding | +7.6367% | [+7.3325, +7.9554]% |

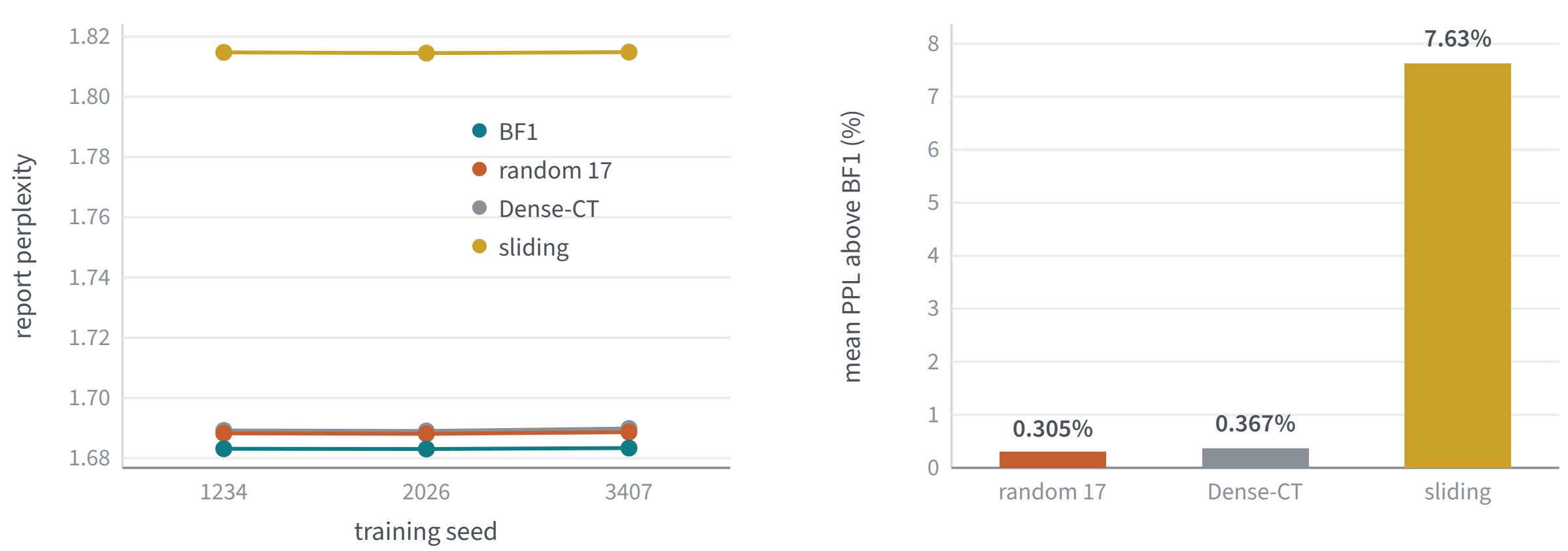

**Figure 5: Stage A report perplexity.** Left: absolute values. Right: relative PPL above BF1 makes the small nonlocal differences and the large local-sliding gap visually distinct.

# 7 Discussion

## 7.1 What the experiments establish

The combined result is stronger than a kernel microbenchmark. BF1 has a public causal route, a proved per-layer interaction law, correctness-gated compiled execution, a long-context crossover against native dense SDPA, a measured partial-model TTFT benefit, and a matched adaptation study that reproduces across training seeds.

The controls also prevent several stronger but unsupported interpretations. Most of the improvement over the unadapted base comes from continued ORCA adaptation. Static-random nonlocal attention remains close to BF1. The large topology effect is the separation between nonlocal routes and local-only sliding; the BF1-specific increment is smaller.

### 7.2 Operator-level versus model-level efficiency

BF1 changes the asymptotic interaction law only in converted layers. The present eight-layer model remains $O(n^2)$ because 20 layers are dense, and the exact historical KV store remains $O(n)$. The all-layer timing curve is therefore a systems projection, not an asymptotic or model-quality result. A complete subquadratic model would require bounding every global layer and measuring all auxiliary costs.

The decode study makes the same accounting point at a smaller scale. Selected-page arithmetic can be very fast while an asymptotically small planning step dominates end-to-end latency. Algorithmic savings must be reported together with route construction, planning, page fetch, synchronization, and framework overhead.

### 7.3 Where BF1 is most immediately useful

The current evidence supports BF1 as a selective retrofit primitive for long-context prefill and for layers or workloads where deterministic local/global coverage and regular physical page access are valuable. Additional studies are needed to determine broader model coverage, graph-draw robustness, transfer to other model scales and corpora, and persistent-plan cached generation.

## 8 Limitations and Reproducibility Boundary

- The study uses one model scale, one primary GPU architecture, batch size one for prefill/TTFT, and one ORCA-derived training family.
- Static-random graph seed 17 is fixed. Training-seed replication does not characterize variation over random graph construction.

- The report split is held out from training and checkpoint selection, but the data family was observed during pilot development. A new external blind corpus remains important.
- Stage A converts only eight layers; the all-layer condition is timing-only and the complete model remains $O(n^2)$.
- BF1 retains an O(n) exact KV store. It is a sparse-read method, not a compressed-memory result.
- The paged-backend execution number is a residual estimate from effective latency minus separately measured planning, not a direct full-generation measurement with persistent plans.
- This paper does not report capability-level retrieval, aggregation, or state-tracking evaluation and makes no claim of universal capability preservation.
- The source archive includes the public route, formal statements, result CSVs, and deterministic figure-generation code. Private checkpoints, conversion tooling, low-level compiler tuning, and proprietary training infrastructure are not released in this version.

## 9 Conclusion

BF1 demonstrates that a causal dyadic block-sparse route can be both algorithmically sparse and practically fast. The operator crosses dense attention at long context, reaches a 10.91× per-layer 32K speedup, and yields a 15.3% whole-model TTFT reduction when applied to eight Qwen3-0.6B layers. Under a matched three-seed adaptation protocol, BF1 consistently achieves the lowest report perplexity, while nonlocal sparse routing decisively outperforms equal-budget local sliding.

BF1 should therefore be interpreted as a validated deterministic sparse-attention operator and selective retrofit primitive with measurable long-context systems value. This paper does not establish a fully sparse model, compressed total KV memory, or broad capability preservation. Those are separate questions requiring their own interventions and evidence.